\documentclass[letterpaper]{article} 
\usepackage{aaai23}  
\usepackage{times}  
\usepackage{helvet}  
\usepackage{courier}  
\usepackage[hyphens]{url}  
\usepackage{graphicx} 
\usepackage{natbib}  
\usepackage{caption} 
\usepackage{bm}
\usepackage{amsmath,amssymb}
\usepackage{algorithm}
\usepackage{algorithmicx}
\usepackage{algpseudocode}
\usepackage{multirow}
\usepackage{booktabs}

\usepackage{newfloat}
\usepackage{listings}
\DeclareCaptionStyle{ruled}{labelfont=normalfont,labelsep=colon,strut=off} 
\floatstyle{ruled}
\newfloat{listing}{tb}{lst}{}
\floatname{listing}{Listing}
\title{ShiftDDPMs: Exploring Conditional Diffusion Models \\ by Shifting Diffusion Trajectories}
\author {
    Zijian Zhang\textsuperscript{\rm 1},
    Zhou Zhao\textsuperscript{\rm 1}\thanks{Corresponding author.},
    Jun Yu\textsuperscript{\rm 2},
    Qi Tian\textsuperscript{\rm 3}
}
\affiliations {
    \textsuperscript{\rm 1} Department of Computer Science and Technology, Zhejiang University\\
    \textsuperscript{\rm 2} School of Computer Science and Technology, Hangzhou Dianzi University\\
    \textsuperscript{\rm 3} Huawei Cloud \& AI\\
    ckczzj@zju.edu.cn, zhaozhou@zju.edu.cn, yujun@hdu.edu.cn, tian.qi1@huawei.com
}

\begin{document}

\maketitle

\begin{abstract}
Diffusion models have recently exhibited remarkable abilities to synthesize striking image samples since the introduction of denoising diffusion probabilistic models (DDPMs).
Their key idea is to disrupt images into noise through a fixed forward process and learn its reverse process to generate samples from noise in a denoising way.
For conditional DDPMs, most existing practices relate conditions only to the reverse process and fit it to the reversal of unconditional forward process.
We find this will limit the condition modeling and generation in a small time window.
In this paper, we propose a novel and flexible conditional diffusion model by introducing conditions into the forward process.
We utilize extra latent space to allocate an exclusive diffusion trajectory for each condition based on some shifting rules, which will disperse condition modeling to all timesteps and improve the learning capacity of model.
We formulate our method, which we call ShiftDDPMs, and provide a unified point of view on existing related methods.
Extensive qualitative and quantitative experiments on image synthesis demonstrate the feasibility and effectiveness of ShiftDDPMs.
\end{abstract}

\section{Introduction and Motivation}
Deep generative models such as Generative Adversarial Networks (GANs)~\cite{goodfellow2014generative}, Variational Autoencoders (VAEs)~\cite{kingma2013auto}, autoregressive models~\cite{van2016pixel} and normalizing flows~\cite{rezende2015variational} have shown remarkable abilities to model complex data distributions and synthesize high-quality samples in various fields.
Diffusion models~\cite{sohl2015deep} are recently brought back into focus by denoising diffusion probabilistic models (DDPMs)~\cite{ho2020denoising}, which exhibits competitive image synthesis results and has been applied in a wide range of data modalities.

\begin{figure}[t]
    \centering
        \includegraphics[width=0.95\columnwidth]{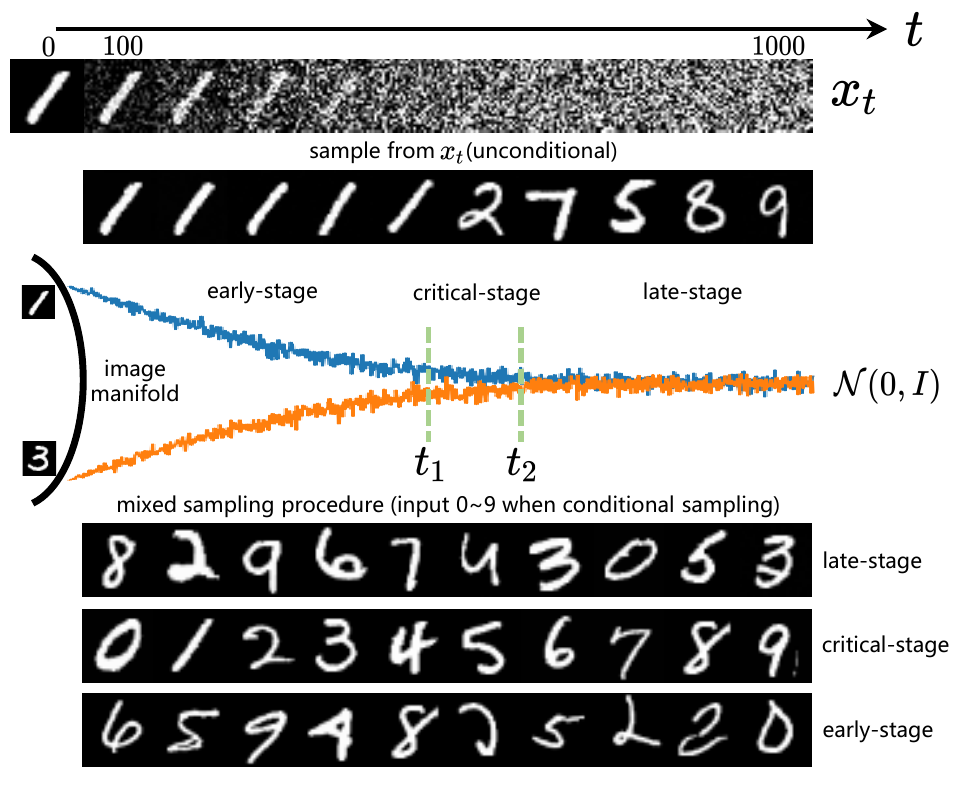}
    \centering
    \caption{Exploration of the mechanism of conditional DDPMs. We grid $1000$ timesteps with a step size of $50$ and perform grid-search for $(t_{1}, t_{2})$ paris to find the shortest critical-stage that can ensure high accuracy of conditional generation. For MNIST, it is $(400, 600)$.}
    \label{fig:introduction}
\end{figure}

Generally, DDPMs gradually disrupt images by adding noise through a fixed forward process and learn its reverse process to generate samples from noise in a denoising way.
There are two main methods to achieve conditional DDPMs.
One is to learn an estimator that can compute the similarity between conditions and noisy data and use it to guide pre-trained unconditional DDPMs to sample towards specified conditions~\cite{dhariwal2021diffusion}.
Another is to train a conditional DDPM from scratch by incorporating conditions into the function approximator of the reverse process.
Both methods try to fit their conditional reverse process to the reversal of fixed unconditional forward process.
This brings up a question: \textit{Can we design a more effective forward process utilizing given conditions to form a new type of conditional DDPMs and benefit from it?}

We investigate this question by exploring the mechanism of how conditional DDPMs achieve conditional sampling based on unconditional forward process, similar to that in PDAE~\cite{zhang2022unsupervised}.
We conduct some experiments, shown in Figure~\ref{fig:introduction}.
Concretely, we train an unconditional DDPM and a conditional one on MNIST~\cite{lecun1998gradient}, respectively.
The conditional one incorporates class labels (one-hot vector) into the function approximator of parameterized reverse process.
The top two rows respectively show the latents $\bm{x}_{t}$ sampled from $\bm{x}_{0}$ for various $t$ and the samples generated by the unconditional DDPM starting from corresponding latents.
Intuitively, the latents for smaller $t$ preserve more high-level information (such as class) of corresponding data, and they will be totally lost when $t$ is large enough.
It means that the diffusion trajectories originating from different data will get entangled, and the latents will become indistinguishable when $t$ is large.
We then divide the diffusion trajectories into three stages: early-stage ($0 \sim t_{1}$), critical-stage ($t_{1} \sim t_{2}$) and late-stage ($t_{2} \sim T$).
Then we design a mixed sampling procedure that employs unconditional sampling but switches to conditional sampling during the specified stage.
Note that the unconditional and conditional reverse process can be connected because they are trained to approximate the same forward process so that they recognize the same pattern of latents.
The bottom three rows show the samples generated by three different mixed sampling procedures, where each row only employs conditional sampling for the right stage.
As we can see, only the samples conditioned on input labels during critical-stage match the input class labels.

These phenomena show that, for unconditional forward process, the key to achieve conditional sampling is to shift and separate the generative trajectories of different conditions during critical-stage.
Besides, to some extent, the training and sampling during early and late stages are independent of conditions and leave the condition modeling and generation to the limited critical-stage.
If we can utilize extra latent space and allocate an exclusive diffusion trajectory for each condition to make the trajectories of different conditions disentangled all the time, it will disperse condition modeling to all timesteps and may improve the learning capacity of model.

Recently, Grad-TTS~\cite{popov2021grad} and PriorGrad~\cite{lee2021priorgrad} introduce conditional forward process with data-dependent priors for audio diffusion models and enable more efficient training than those with unconditional forward process.
However, their differences and connections have not been discussed, and there has not been a comprehensive exploration of this kind of methods, especially for image diffusion models.
In this work, we systematically study how to design controllable diffusion trajectories according to conditions and its effect for conditional diffusion models.
Our main contributions contain:
\begin{itemize}
    \item We systemically introduce conditional forward process for diffusion models and provide a unified point of view on existing related approaches.
    \item By shifting diffusion trajectories, ShiftDDPMs improve the utilization rate of latent space and the learning capacity of model.
    \item We demonstrate the feasibility and effectiveness of ShiftDDPMs on various image synthesis tasks with extensive experiments.
\end{itemize}

\section{Related Works}
Diffusion models~\cite{sohl2015deep,ho2020denoising} are an emerging family of generative models and have exhibited remarkable abilities to synthesize high-quality samples.
Numerous studies~\cite{song2020score,song2020denoising,dhariwal2021diffusion,liu2022pseudo} and applications~\cite{chen2020wavegrad,saharia2022photorealistic,huang2022fastdiff,huang2022prodiff,ye2022syntaspeech,ye2023geneface} have further improved and expanded diffusion models.
Among existing practices of conditional diffusion models, only Grad-TTS~\cite{popov2021grad} and PriorGrad~\cite{lee2021priorgrad} involve conditions in forward process but, nonetheless, they are totally different methods.
We will demonstrate their differences under the point of view of ShiftDDPMs.

\section{ShiftDDPMs}
\subsection{Background}
DDPMs~\cite{ho2020denoising} employ a forward process that sequentially destroys data distribution $q({\bm{x}_{0}})$ into $\mathcal{N}(\bm{0},\bm{I})$ with Markov diffusion kernels defined by a fixed variance schedule $\{\beta_{t}\}_{t=1}^{T}$:
\begin{equation}\label{original_forward_step}
    \begin{aligned}
        q(\bm{x}_{t} | \bm{x}_{t-1}) = \mathcal{N} (\sqrt{1-\beta_{t}}\bm{x}_{t-1} , \beta_{t}\bm{I}) ,
    \end{aligned}
\end{equation}
which admits sampling $\bm{x}_{t}$ from $\bm{x}_{0}$ for any timestep $t$ in closed form:
\begin{equation}\label{original_forward_direct}
    \begin{aligned}
        q(\bm{x}_{t} | \bm{x}_{0}) = \mathcal{N} (\sqrt{\bar{\alpha}_{t}}\bm{x}_{0} , (1-\bar{\alpha}_{t})\bm{I}) .
    \end{aligned}
\end{equation}
Then a parameterized Markov chain is trained to fit the reversal of forward process, denoising an arbitrary Gaussian noise to a data sample:
\begin{equation}
    \begin{aligned}
        p_{\theta}(\bm{x}_{t-1} | \bm{x}_{t}) = \mathcal{N} (\bm{\mu}_{\theta}(\bm{x}_{t},t) , \bm{\Sigma}_{\theta}(\bm{x}_{t},t)) .
    \end{aligned}
\end{equation}
Training is performed by maximizing the model log likelihood with some parameterization and simplication:
\begin{equation}
    \begin{aligned}
        L(\theta) = \mathbb{E}_{t,\bm{x}_{0},\epsilon}\bigg[ \| \epsilon - \epsilon_{\theta}(\sqrt{\bar{\alpha}_{t}}\bm{x}_{0}+\sqrt{1-\bar{\alpha}_{t}}\epsilon, t) \|^{2} \bigg] .
    \end{aligned}
\end{equation}
See Appendix A for full details of DDPMs.

\subsection{Conditional Forward Process}
We aim to shift the diffusion trajectories in some way related to conditions.
An intuitive way is to directly rewrite the Gaussian distribution in Eq.(\ref{original_forward_direct}) as:
\begin{equation}
    \begin{aligned}
        q(\bm{x}_{t} | \bm{x}_{0}, \bm{c}) = \mathcal{N} (\sqrt{\bar{\alpha}_{t}}\bm{x}_{0} + k_{t} \cdot \bm{E}(\bm{c}) , (1-\bar{\alpha}_{t}) \bm{\Sigma}(\bm{c})) .
    \end{aligned}
\end{equation}
Specifically, $k_{t} \cdot \bm{E}(\bm{c})$ is the cumulative mean shift of diffusion trajectories at $t$-th step, where $k_{t}$ is a shift coefficient schedule that decides the shift mode and $\bm{E}(\cdot)$ is a function which we call shift predictor that maps conditions into the latent space.
$\bm{\Sigma}(\bm{c})$ is a diagonal covariance matrix, where $\bm{\Sigma}(\cdot)$ is some function similar to $\bm{E}(\cdot)$.
Comparing the diffusion trajectories to water pipes, then $k_{t} \cdot \bm{E}(\bm{c})$ is employed to change their directions and $\bm{\Sigma}(\bm{c})$ is employed to change their size in latent space.
Note that both $\bm{E}(\cdot)$ and $\bm{\Sigma}(\cdot)$ can be fixed or trainable.
In our experiments on image synthesis, trainable $\bm{\Sigma}(\cdot)$ leads to complex training and sampling procedure, unstable training and poor results, so we fix $\bm{\Sigma}(\bm{c}) = \bm{I}$ like that in Eq.(\ref{original_forward_direct}).
For generalization, we still use $\bm{\Sigma}(\bm{c})$ in our derivations.
For simplicity, we use following substitution:
\begin{equation}\label{our_forward_direct}
    \begin{aligned}
        q(\bm{x}_{t} | \bm{x}_{0}, \bm{c}) = \mathcal{N} (\sqrt{\bar{\alpha}_{t}}\bm{x}_{0} + \bm{s}_{t} , (1-\bar{\alpha}_{t}) \bm{\Sigma}) ,
    \end{aligned}
\end{equation}
where $\bm{s}_{t} = k_{t} \cdot \bm{E}(\bm{c})$ and $\bm{\Sigma} = \bm{\Sigma}(\bm{c})$.
We will discuss how to choose $k_{t}$ and $\bm{E}(\cdot)$ in later sections.

With Eq.(\ref{our_forward_direct}), we can derive corresponding forward diffusion kernels (See proof in Appendix A):
\begin{equation}\label{our_forward_step}
    \begin{aligned}
        q(\bm{x}_{t} | \bm{x}_{t-1}, \bm{c}) = \mathcal{N} (\sqrt{\alpha_{t}}\bm{x}_{t-1} + \bm{s}_{t} - \sqrt{\alpha_{t}}\bm{s}_{t-1} , \beta_{t}\bm{\Sigma}) ,
    \end{aligned}
\end{equation}
where $\bm{s}_{0} = \bm{0}$ (i.e. $k_{0}=0$).
Intuitively, our forward diffusion kernels introduce a small perturbation conditioned on $\bm{c}$ to original ones shown in Eq.(\ref{original_forward_step}).

With Eq.(\ref{our_forward_direct}) and Eq.(\ref{our_forward_step}), the posterior distributions of forward steps for $t > 1$ can be derived from Bayes' rule (See proof in Appendix A):
\begin{equation}\label{our_forward_posterior}
    \begin{aligned}
        & q(\bm{x}_{t-1}|\bm{x}_{t},\bm{x}_{0}, \bm{c}) = \mathcal{N} (\frac{\sqrt{\bar{\alpha}_{t-1}}\beta_{t}}{1-\bar{\alpha}_{t}}\bm{x}_{0} + \frac{\sqrt{\alpha_{t}}(1-\bar{\alpha}_{t-1})}{1-\bar{\alpha}_{t}}\bm{x}_{t} \\
        &  - \frac{\sqrt{\alpha_{t}}(1-\bar{\alpha}_{t-1})}{1-\bar{\alpha}_{t}}\bm{s}_{t} + \bm{s}_{t-1} \,, \, \frac{1-\bar{\alpha}_{t-1}}{1-\bar{\alpha}_{t}}\beta_{t} \bm{\Sigma}) \,.
    \end{aligned}
\end{equation}

\subsection{Parameterized Reverse Process}
The reverse process starts at $p(\bm{x}_{T}) = \mathcal{N}(\bm{s}_{T}, \bm{\Sigma})$, which is an approximation of $q(\bm{x}_{T} | \bm{x}_{0}, \bm{c})$, and employs parameterized kernels $p_{\theta}(\bm{x}_{t-1} | \bm{x}_{t}, \bm{c})$ to fit $q(\bm{x}_{t-1}|\bm{x}_{t},\bm{x}_{0}, \bm{c})$.

According to Eq.(\ref{our_forward_direct}), $\bm{x}_{0}$ can be represented as:
\begin{equation}\label{parameterization}
    \begin{aligned}
        \bm{x}_{0} = \frac{1}{\sqrt{\bar{\alpha}_{t}}} \big( \bm{x}_{t} - \bm{s}_{t} - \sqrt{1-\bar{\alpha}_{t}}\bm{\epsilon} \big) \,,
    \end{aligned}
\end{equation}
where $\bm{\epsilon} \sim \mathcal{N}(\bm{0},\bm{\Sigma})$.
Then we take it into Eq.(\ref{our_forward_posterior}) and derive the posterior mean of forward steps:
\begin{equation}\label{new_our_forward_posterior}
    \begin{aligned}
        \mathbb{E}\big[q(\bm{x}_{t-1}|\bm{x}_{t},\bm{x}_{0}, \bm{c})\big] =& \frac{1}{\sqrt{\alpha_{t}}}(\bm{x}_{t} - \frac{\beta_{t}}{\sqrt{1-\bar{\alpha}_{t}}}\bm{\epsilon}) \\
        &- \frac{1}{\sqrt{\alpha_{t}}}\bm{s}_{t} + \bm{s}_{t-1} \,,
    \end{aligned}
\end{equation}
where all things are available except $\bm{\epsilon}$.
We can employ a model $\bm{\epsilon}_{\theta}(\bm{x}_{t}, t)$ to predict $\bm{\epsilon}$.
Note that there is no need to feed $\bm{c}$ into $\bm{\epsilon}_{\theta}$ because we have encoded it into condition-dependent trajectories (i.e., in $\bm{x}_{t}$) so that the model does not need its guidance.

Further improvements come from another parameterization because $\bm{\epsilon}$ in Eq.(\ref{parameterization}) is given by:
\begin{equation}
    \begin{aligned}
        \bm{\epsilon} = \frac{\bm{x}_{t} - \sqrt{\bar{\alpha}_{t}}\bm{x}_{0}}{\sqrt{1-\bar{\alpha}_{t}}} - \frac{\bm{s}_{t}}{\sqrt{1-\bar{\alpha}_{t}}} \,,
    \end{aligned}
\end{equation}
where the second term is available.
Therefore we can employ a model $\bm{g}_{\theta}(\bm{x}_{t}, t)$ to predict the first term for training.
We find this parameterization achieves better performance than predicting $\bm{\epsilon}$ directly.
Then we can get the predicted posterior distributions parameterized by $\theta$:
\begin{equation}
    \begin{aligned}
        p_{\theta}(\bm{x}_{t-1} | \bm{x}_{t}, \bm{c}) = \mathcal{N} (\frac{1}{\sqrt{\alpha_{t}}}\bigg[\bm{x}_{t} - \frac{\beta_{t}}{\sqrt{1-\bar{\alpha}_{t}}}\bm{g}_{\theta}(\bm{x}_{t}, t)\bigg] \\
        - \frac{\sqrt{\alpha_{t}}(1-\bar{\alpha}_{t-1})}{1-\bar{\alpha}_{t}}\bm{s}_{t} + \bm{s}_{t-1}  ,  \frac{1-\bar{\alpha}_{t-1}}{1-\bar{\alpha}_{t}}\beta_{t} \bm{\Sigma}) .
    \end{aligned}
\end{equation}

\begin{algorithm}[t]
    \caption{Training} \label{alg:training}
    \small
    \begin{algorithmic}[1]
    \Repeat
        \State $\bm{x}_0, \bm{c} \sim q(\bm{x}_0)$
        \State $t \sim \mathrm{Uniform}(\{1, \dotsc, T\})$
        \State $\bm{s}_{t}= k_{t} \cdot \bm{E}(\bm{c})\,, \ \ \bm{\Sigma} = \bm{\Sigma}(\bm{c})$
        \State $\bm{\epsilon}\sim\mathcal{N}(\bm{0},\bm{\Sigma})$
        \State $\bm{x}_{t} = \sqrt{\bar{\alpha}_{t}}\bm{x}_{0} + \bm{s}_{t} + \sqrt{1-\bar{\alpha}_{t}}\bm{\epsilon}$
        \State Optimize $\| \frac{\bm{x}_{t} - \sqrt{\bar{\alpha}_{t}}\bm{x}_{0}}{\sqrt{1-\bar{\alpha}_{t}}} - \bm{g}_{\theta}(\bm{x}_{t}, t) \|^{2}_{\bm{\Sigma}^{-1}}$
    \Until{converged}
    \end{algorithmic}
\end{algorithm}

\begin{algorithm}[t]
    \caption{Sampling} \label{alg:sampling}
    \small
    \begin{algorithmic}[1]
    \State $\bm{s}_{T} = k_{T} \cdot \bm{E}(\bm{c})\,, \ \ \bm{\Sigma} = \bm{\Sigma}(\bm{c})$ 
    \State $\bm{x}_T \sim \mathcal{N}(\bm{s}_{T}, \bm{\Sigma})$
    \For{$t=T, \dotsc, 1$}
        \State $\bm{z} \sim \mathcal{N}(\bm{0}, \bm{\Sigma})$ if $t > 1$, else $\bm{z} = \bm{0}$
        \State $\bm{x}_{t-1} = \frac{1}{\sqrt{\alpha_{t}}}\big[\bm{x}_{t} - \frac{\beta_{t}}{\sqrt{1-\bar{\alpha}_{t}}}\bm{g}_{\theta}(\bm{x}_{t}, t)\big]$ 
        \Statex $ \,\,\,\,\,\,\,\,\,\,\,\,\,\,\,\,\,\,\,\,\,\,\,\,\,\,\,\,\,\, - \frac{\sqrt{\alpha_{t}}(1-\bar{\alpha}_{t-1})}{1-\bar{\alpha}_{t}}\bm{s}_{t} + \bm{s}_{t-1} + \sqrt{\frac{1-\bar{\alpha}_{t-1}}{1-\bar{\alpha}_{t}}\beta_{t}}\bm{z}$
    \EndFor
    \State \textbf{return} $\bm{x}_0$
    \end{algorithmic}
\end{algorithm}

\subsection{Training Objective}
With our conditional forward process and corresponding reverse process, our training objective can be represented as (See proof in Appendix A):
\begin{equation}
    \begin{aligned}
        L = c + \sum_{t=1}^{T} \gamma_{t} \mathbb{E}_{\bm{x}_{0},\epsilon}\bigg[ \| \frac{\bm{x}_{t} - \sqrt{\bar{\alpha}_{t}}\bm{x}_{0}}{\sqrt{1-\bar{\alpha}_{t}}} - \bm{g}_{\theta}(\bm{x}_{t}, t) \|^{2}_{\bm{\Sigma}^{-1}} \bigg] ,
    \end{aligned}
\end{equation}
where $c$ is some constant, $\bm{x}_{0} \sim q(\bm{x_{0}})$, $\bm{\epsilon} \sim \mathcal{N}(\bm{0},\bm{\Sigma})$, $\bm{x}_{t} = \sqrt{\bar{\alpha}_{t}}\bm{x}_{0} + \bm{s}_{t} + \sqrt{1-\bar{\alpha}_{t}}\bm{\epsilon}$, $\|\bm{x}\|^{2}_{\bm{\Sigma}^{-1}} = \bm{x}^{T}\bm{\Sigma}^{-1}\bm{x}$, $\gamma_{1}=\frac{1}{2\alpha_{1}}$ and $\gamma_{t}=\frac{\beta_{t}}{2\alpha_{t}(1-\bar{\alpha}_{t-1})}$ for $t \ge 2$.
During training, we follow DDPMs~\cite{ho2020denoising} to adopt the simplified training objective by uniformly sampling $t$ between $1$ and $T$ and ignoring loss weight $\gamma_{t}$.
Algorithm~\ref{alg:training} and Algorithm~\ref{alg:sampling} describe our training and sampling procedure.
Note that $\bm{E}(\cdot)$ and $\bm{\Sigma}(\cdot)$ will be optimized along with $\theta$ if they are trainable.

\subsection{Intuitive Interpretation}
Assume that $\bm{\Sigma} = \bm{I}$ and $\bm{x}^{\prime}_{t} = \sqrt{\bar{\alpha}_{t}}\bm{x}_{0} + \sqrt{1-\bar{\alpha}_{t}}\bm{\epsilon}$, DDPMs employ $\bm{\epsilon}_{\theta}(\bm{x}^{\prime}_{t}, \bm{c}, t)$ to predict $\epsilon = \frac{\bm{x}^{\prime}_{t} - \sqrt{\bar{\alpha}_{t}}\bm{x}_{0}}{\sqrt{1-\bar{\alpha}_{t}}}$, while ShiftDDPMs employ $\bm{g}_{\theta}(\bm{x}^{\prime}_{t} + \bm{s}_{t}, t)$ to predict $\frac{\bm{x}^{\prime}_{t} + \bm{s}_{t} - \sqrt{\bar{\alpha}_{t}}\bm{x}_{0}}{\sqrt{1-\bar{\alpha}_{t}}}$.
They are trained to predict the same pattern of objective but with different input (i.e. $\frac{\text{input} - \sqrt{\bar{\alpha}_{t}}\bm{x}_{0}}{\sqrt{1-\bar{\alpha}_{t}}}$).
Compared with DDPMs, ShiftDDPMs transfer input condition $\bm{c}$ onto diffusion trajectories by shifting $\bm{x}^{\prime}_{t}$ to $\bm{x}^{\prime}_{t}+\bm{s}_{t}$, which allows conditional training and sampling without feeding $\bm{c}$ into the network.
For DDPMs, only the training and sampling during limited critical-stage plays a key role for condition modeling and generation, while ShiftDDPMs disperse it to all timesteps and improve the utilization rate of latent space, which may lead to a better performance.

Furthermore, if $\bm{E}(\cdot)$ is trainable, it will be optimized to find an optimal shift in latent space to specialize the diffusion trajectories of different conditions and make them disentangle as much as possible.
The term $ \bm{d}_{t} = -\frac{1}{\sqrt{\alpha_{t}}}\bm{s}_{t} + \bm{s}_{t-1}$ in Eq.(\ref{new_our_forward_posterior}) will amend the sampling trajectories in every step to ensure they can finally fall on the data manifold.

Next, we will show that the forward process of Grad-TTS~\cite{popov2021grad} and PriorGrad~\cite{lee2021priorgrad} correspond to a special choice of $k_{t}$, respectively.

\subsection{Prior-Shift}
Grad-TTS~\cite{popov2021grad} proposes a score-based text-to-speech generative model with the prior mean predicted by text encoder and aligner.
Specifically, it defines a forward process satisfying the following SDE:
\begin{equation}\label{gradtts}
    \begin{aligned}
        \mathrm{d} \bm{X}_{t} = \frac{1}{2} (\bm{\mu} - \bm{X}_{t})\beta_{t}\mathrm{d}t + \sqrt{\beta_{t}}\mathrm{d}\bm{W}_{t} \,,
    \end{aligned}
\end{equation}
where $\bm{\mu}$ corresponds to $\bm{E}(\bm{c})$ of our system ($\bm{E}(\cdot)$ represents the parameterized text encoder and aligner, $\bm{c}$ represents the input text).
We show that $k_{t} = 1 - \sqrt{\bar{\alpha}_{t}}$ match a discretization of Eq.(\ref{gradtts}) (See proof in Appendix A).
For forward process, $k_{t}$ increases from $0$ to $1$ and leads $\bm{x}_{t}$ to shift to $\bm{\mu}$ as $t$ increases.
For reverse process, we have:
\begin{equation}
    \begin{aligned}
        &\bm{d}_{t} = (1- \frac{1}{\sqrt{\alpha}_{t}})\bm{\mu}\,,
    \end{aligned}
\end{equation}
where $1- \frac{1}{\sqrt{\alpha}_{t}} < 0$ because the reverse process starts from $\mathcal{N} (\bm{\mu}, \bm{I})$ and it needs to eliminate the cumulative shift $\bm{\mu}$ of forward process.
From the view of diffusion trajectories, Grad-TTS changes the ending point of trajectories, so we name the shift mode as Prior-Shift.

Note that Grad-TTS still takes $\bm{\mu}$ as an additional input to the score estimator, but we have stated that it is unnecessary.
However, doing this will get at least not worse results, but also introduces additional parameter and computation.

\subsection{Data-Normalization}
PriorGrad~\cite{lee2021priorgrad} employs a forward process as follows:
\begin{equation}\label{priorgrad}
    \begin{aligned}
        \bm{x}_{t} = \sqrt{\bar{\alpha}_{t}}(\bm{x}_{0} - \bm{\mu}) + \sqrt{1-\bar{\alpha}_{t}}\bm{\epsilon} \,,
    \end{aligned}
\end{equation}
where $\bm{\epsilon} \sim \mathcal{N} (\bm{0}, \bm{\Sigma})$.
Obviously, $k_{t} = - \sqrt{\bar{\alpha}_{t}}$ satisfies Eq.(\ref{priorgrad}).
For forward process, it first normalizes $\bm{x}_{0}$ by subtracting its corresponding prior mean $\bm{\mu}$ and then trains a diffusion model on normalized $\bm{x}_{0}$ with prior $\mathcal{N} (\bm{0}, \bm{\Sigma})$.
For reverse process, we have:
\begin{equation}
    \begin{aligned}
        \bm{d}_{1} = \bm{\mu} \,,\,\, \bm{d}_{t>1} = \bm{0} .
    \end{aligned}
\end{equation}
Intuitively, the reverse process starts from $\mathcal{N} (\bm{0}, \bm{\Sigma})$ and has no amendments all the time except the last step, where it adds prior mean $\bm{\mu}$ to the output (denormalization).
From the view of diffusion trajectories, PriorGrad resets the starting point of trajectories on the data manifold, so we name the shift mode as Data-Normalization.

Unlike Prior-Shift that disperses the cumulative shift to all points on the diffusion trajectories, Data-Normalization does not disentangle the diffusion trajectories so that it must feed $\bm{c}$ into the network to guide sampling.
However, by carefully designing $\bm{\Sigma}$, it can achieve the same precision with a simpler network and have a faster convergence rate under some constraints~\cite{lee2021priorgrad}.
Data-Normalization is more suitable for variance-sensitive data such as audio.

\subsection{Quadratic-Shift}
Except for Prior-Shift, we propose a shift mode to disentangle the diffusion trajectories of different conditions by making the concave trajectories shown in Figure~\ref{fig:introduction} convex.
In this case, we don't change their starting or ending point, and $\bm{E}(\bm{c})$ becomes a middle point, where they first progress to it and then go away from it.
Therefore $k_{t}$ should be similar to some quadratic function opening downwards with $k_{1} \approx 0$ and $k_{T} \approx 0$.
Empirically, we choose $k_{t} = \sqrt{\bar{\alpha}_{t}} (1-\sqrt{\bar{\alpha}_{t}})$.
We name the shift mode as Quadratic-Shift.

\section{Experiments}
In this section, we conduct several conditional image synthesis experiments with ShiftDDPMs.
Note that we always set $\bm{\Sigma}(\bm{c}) = \bm{I}$.
Full implementation details of all experiments can be found in Appendix B.

\begin{figure}[t]
    \centering
    \includegraphics[width=1.0\linewidth]{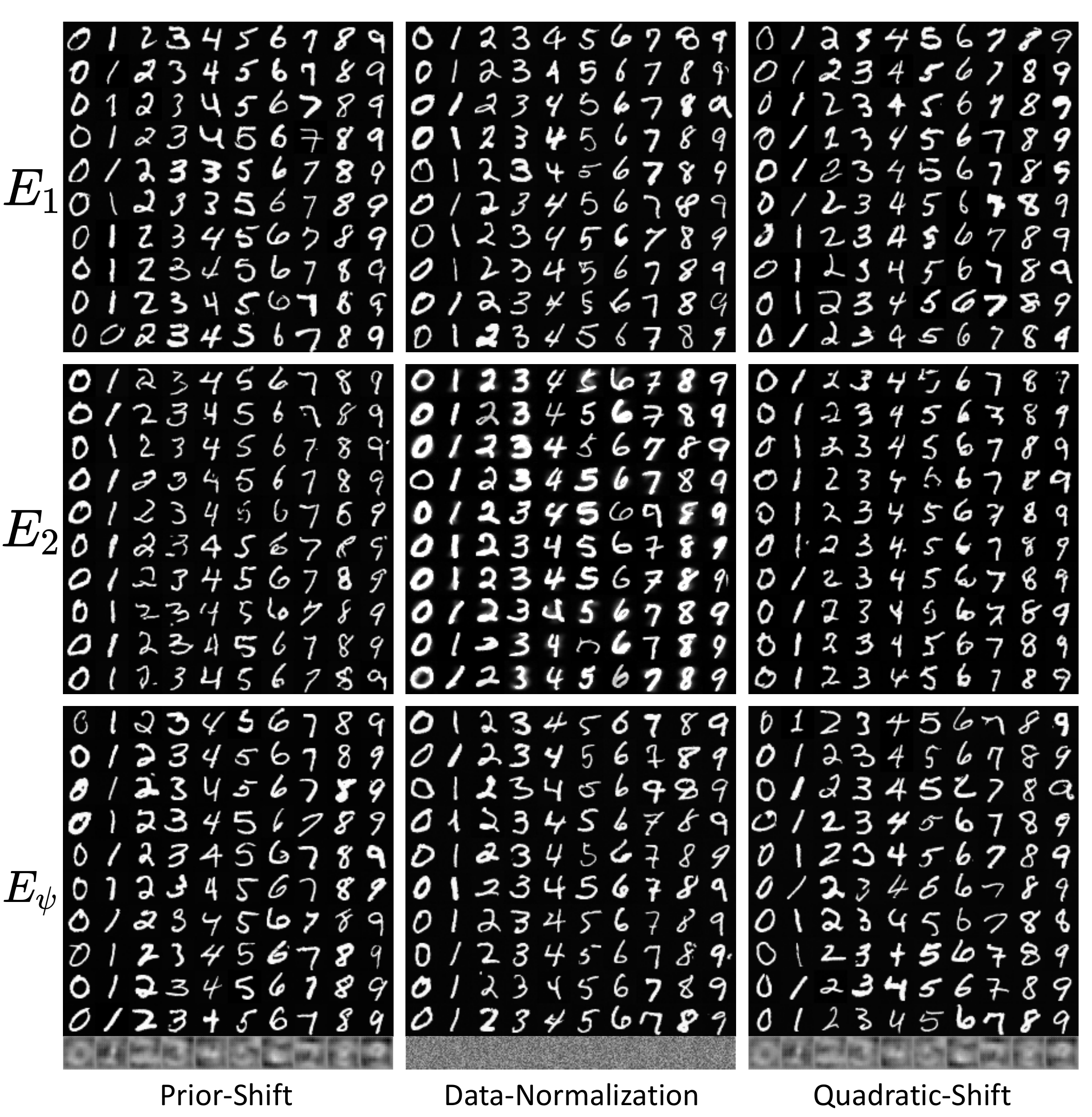}
    \caption{
         $32 \times 32$ conditional MNIST samples for different shift modes with different shift predictors. The last row visualize the learned $\bm{E}_{\psi}(\cdot)$.
    }
    \label{fig:mnist}
\end{figure}

\subsection{Effectiveness of Conditional Sampling}
We first verify the effectiveness of ShiftDDPMs with three shift modes on toy dataset MNIST~\cite{lecun1998gradient}.
We employ two fixed shift predictors ($\bm{E}_{1}(\cdot)$ and $\bm{E}_{2}(\cdot)$) and a trainable one ($\bm{E}_{\psi}(\cdot)$ with parameters ${\psi}$), mapping a one-hot vector $\bm{c}$ to a $32 \times 32$ matrix.
Specifically, $\bm{E}_{1}(\cdot)$ takes $10$ evenly spaced numbers over $[-1, 1]$ and expands each number into a $32 \times 32$ matrix.
$\bm{E}_{2}(\cdot)$ takes the mean of all training data belonging to the specified class.
$\bm{E}_{\psi}(\cdot)$ employs stacked transposed convolution layers to compute the matrix.

Figure~\ref{fig:mnist} presents the conditional MNIST samples for different shift modes with different shift predictors.
As we can see, all models work for conditional generation, and the visualization of learned $\bm{E}_{\psi}(\bm{c})$ for Prior-Shift and Quadratic-Shift contain the general shape of corresponding class, which means that they learn specialized trajectories for different conditions.
Data-Normalization must feed $\bm{c}$ into the model so it may ignore the shift.

Despite the success of the fixed shift predictor on MNIST, we get poor sample results when modeling complex data distribution such as CIFAR-10.
Therefore we will always employ trainable shift predictor $\bm{E}_{\psi}$ with parameter $\psi$ in the following experiments.

\begin{figure}[t]
    \centering
    \includegraphics[width=1.0\linewidth]{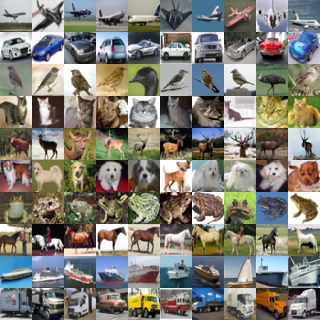}
    \caption{
         $32 \times 32$ conditional CIFAR-10 samples for Quadratic-Shift.
    }
    \label{fig:cifar}
\end{figure}

\begin{table}[t]
        \centering
        \begin{tabular}{lccc}
            \toprule
            Model & IS$\uparrow$ & FID$\downarrow$ & NLL$\downarrow$ \\
            \midrule
            \textbf{Unconditional} \\
            \midrule
            DDPM & $9.46$ & $3.17$ & $\le 3.75$ \\
            our DDPM & $9.52$ & $3.13$ & $\le 3.72$ \\
            \midrule
            \textbf{Conditional} \\
            \midrule
            cond. DDPM & $9.59$ & $3.12$ & $\le 3.74$ \\
            cls. DDPM & $9.17$ & $5.85$ & $-$ \\
            Prior-Shift & $9.54$ & $3.06$ & $\le 3.71$ \\
            cond. Prior-Shift & $9.65$ & $3.06$ & $\le 3.70$ \\
            Data-Normalization & $9.14$ & $5.51$ & $-$ \\
            Quadratic-Shift & $9.67$ & $3.05$ & $\le \bm{3.69}$ \\
            cond. Quadratic-Shift & $\bm{9.74}$ & $\bm{3.02}$ & $\le 3.70$ \\
            \bottomrule
        \end{tabular}
    \caption{Quantitative results of conditional sample quality on CIFAR-10. NLL measured in bits/dim.}
    \label{table:cifar}
\end{table}

\subsection{Sample Quality}
We further evaluate ShiftDDPMs on CIFAR-10~\cite{krizhevsky2009learning}.
For a fair comparison, we retrain a DDPM as baseline (our DDPM) and then use the same experimental settings and resources to train other models.
We train a traditional conditional DDPM (cond. DDPM) by incorporating class labels into the function approximator of reverse process.
Moreover, we train a time-dependent classifier~\cite{sohl2015deep,song2020score,dhariwal2021diffusion} on noisy images and use its gradients to guide (our DDPM) to sample towards specified class (cls. DDPM).
For ShiftDDPMs, we train three models, including Prior-Shift, Data-Normalization, and Quadratic-Shift, all with trainable shift predictors.
Furthermore, we employ another two models (cond. Prior-Shift and cond. Quadratic-Shift) by incorporating class labels into the reverse process of Prior-Shift and Quadratic-Shift, with the same method with (cond. DDPM).
Figure~\ref{fig:cifar} presents some conditional CIFAR-10 samples generated by Quadratic-Shift.
Table~\ref{table:cifar} shows Inception Score, FID, negative log likelihood for these models.

As we can see, our retrained unconditional DDPM is slightly better than the original one with the help of improved settings.
With the help of conditional knowledge, conditional DDPM outperforms unconditional DDPM.
Classifer-guided DDPM has poor results because it is sensitive to the classifier.
Data-Normalization has an unstable training process and poor results, which means that it is not suitable for image synthesis.
Both Prior-Shift and Quadratic-Shift outperform conditional DDPM, which proves that conditional forward process can improve the learning capacity of ShiftDDPMs.
Although incorporating class labels can slightly improve their performance, it also introduces additional computational and parameter complexity.

\subsection{Adaption to DDIM for Fast Sampling}
DDIMs~\cite{song2020denoising} generalize the forward process of DDPMs to non-Markovian process with an equivalent objective for training, which enables us to employ an accelerated reverse process with pre-trained DDPMs.
Fortunately, ShiftDDPMs can be adapted to ShiftDDIMs.
Specifically, we can generate $\bm{x}_{t-1}$ from $\bm{x}_{t}$ via:
\begin{equation}
    \begin{aligned}
        &\bm{x}_{t-1} = \frac{1}{\sqrt{\alpha_{t}}}\big[ \bm{x}_{t} - \sqrt{1 - \bar{\alpha_{t}}} \bm{g}_{\theta}(\bm{x}_{t}, t) \big] + \bm{s}_{t-1} \\ 
        &+ \sqrt{1-\bar{\alpha}_{t-1}- \sigma_{t}^{2}} \cdot \bigg[ \bm{g}_{\theta}(\bm{x}_{t}, t) - \frac{\bm{s}_{t}}{\sqrt{1 - \bar{\alpha}_{t}}} \bigg] + \sigma_{t}\bm{\epsilon}_{t} ,
    \end{aligned}
\end{equation}
where $\bm{\epsilon}_{t} \sim \mathcal{N}(\bm{0}, \bm{\Sigma})$ (See proof in Appendix A).

\begin{figure}[t]
    \centering
    \includegraphics[width=1.0\linewidth]{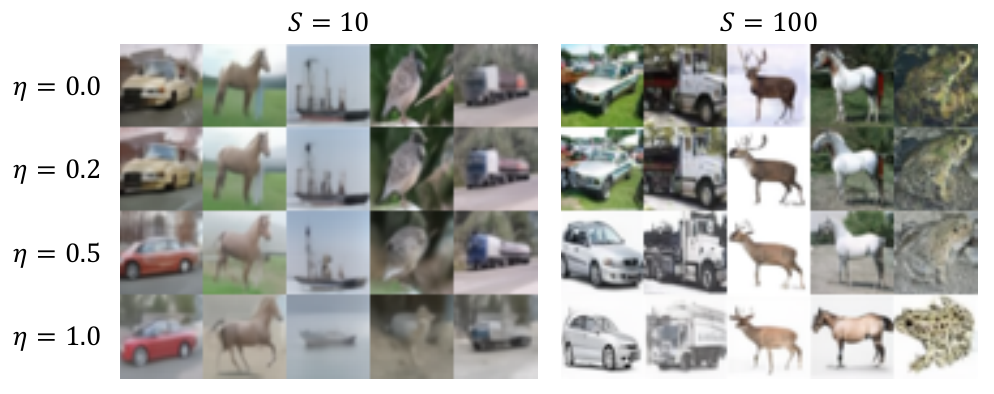}
    \caption{$32 \times 32$ conditional CIFAR-10 samples for Quadratic-Shift. We use fixed input and noise during sampling.}
    \label{fig:ddim}
\end{figure}
\begin{table}
\centering
\begin{tabular}{cc|ccccc}
    \toprule
    \multicolumn{2}{c|}{S} & 10 & 20 & 50 & 100 \\
    \midrule
    \multirow{4}{*}{$\eta$}
    & 0.0 & 14.25 & 7.95 & 5.22 & 3.93 \\
    & 0.2 & 14.16 & 7.88 & 5.30 & 4.06 \\ 
    & 0.5 & 16.96 & 9.12 & 6.18 & 4.43 \\
    & 1.0 & 25.33 & 11.67 & 9.81 & 5.70 \\
    \midrule
    \multicolumn{2}{c|}{$\hat{\sigma}$} & 264.32 & 118.61 & 36.24 & 10.95 \\
    \bottomrule
\end{tabular}
\caption{FID of conditional sample quality on CIFAR-10 for Quadratic-Shift.}
\label{tab:ddim}
\end{table}

Then we employ $\tau = \{\tau_{1}, \cdots, \tau_{S}\}$, which is an increasing sub-sequence of $[1, \cdots, T]$ of length $S$, for accelerated sampling.
The corresponding variance become $\sigma_{\tau_{i}}(\eta) = \eta \sqrt{\frac{1-\bar{\alpha}_{\tau_{i-1}}}{1-\bar{\alpha}_{\tau_{i}}}}\sqrt{1-\frac{\bar{\alpha}_{\tau_{i}}}{\bar{\alpha}_{\tau_{i-1}}}}$, where $\eta$ is a hyperparameter that we can directly control.
Figure~\ref{fig:ddim} and Table~\ref{tab:ddim} presents the conditional CIFAR-10 samples generated by Quadratic-Shift mode and its FID with different sampling steps and $\eta$.
ShiftDDIMs can still keep competitive FID even though it only samples for $100$ steps.

\begin{figure}[t]
        \centering
        \includegraphics[width=2.7in]{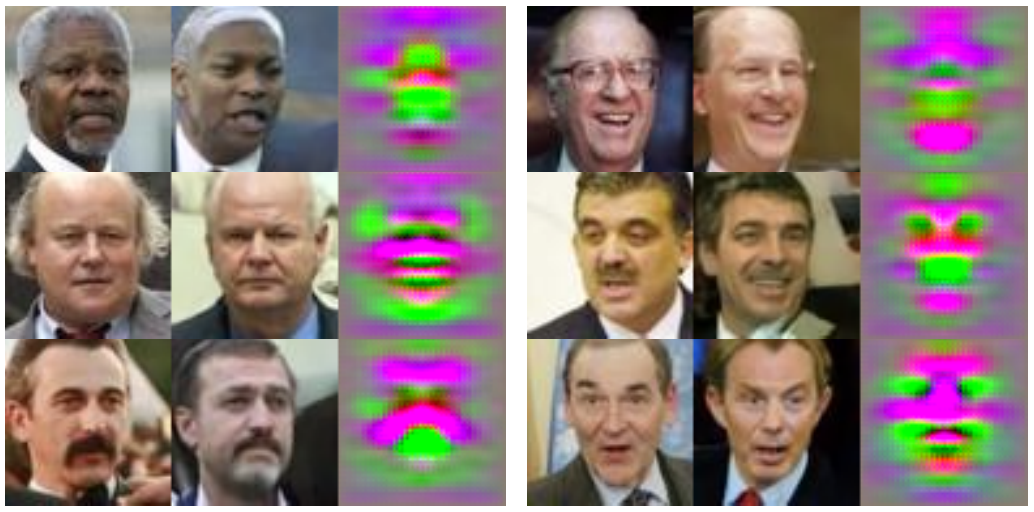}
        \caption{
            $64 \times 64$ conditional LFW samples for Quadratic-Shift. From left to right are ground truth image (from test set), generated image and learned $\bm{E}_{\psi}(\bm{c})$.
        }
        \label{fig:lfw1}
\end{figure}
\begin{figure}[t]
        \centering
        \includegraphics[width=3.2in]{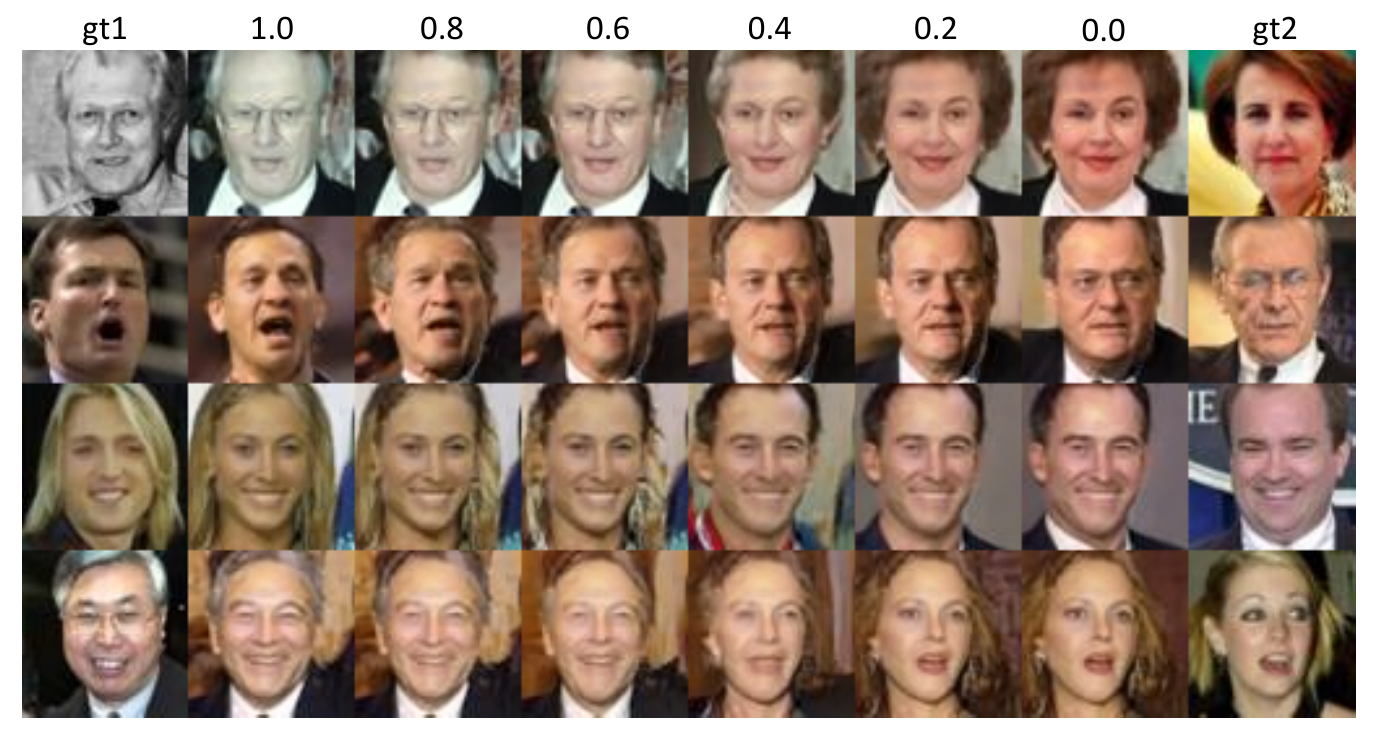}
        \caption{
            $64 \times 64$ conditional LFW interpolations for Quadratic-Shift. We use fixed input and noise during sampling.
        }
        \label{fig:lfw2}
\end{figure}

\subsection{Interpolation of Diffusion Trajectories}
DDPMs~\cite{ho2020denoising} show that one can interpolate the latents of two source data, decode the interpolated latent by the reverse process and get a sample similar to the interpolation of two source data.
Inspired by this phenomenon, we can try to interpolate the diffusion trajectories of different conditions, which is equivalent to interpolating between different $\bm{s}_{t}$, such as $ \hat{\bm{s}}_{t} = \lambda \cdot k_{t} \cdot \bm{E}_{\psi}(\bm{c}_{1}) + (1-\lambda) \cdot k_{t} \cdot \bm{E}_{\psi}(\bm{c}_{2})$ for two different conditions $\bm{c}_{1}$ and $\bm{c}_{2}$.
In theory, $\bm{s}_{t}$ decide the direction of diffusion trajectories and the interpolated $\hat{\bm{s}}_{t}$ will take the median direction, which can lead the reverse process to generate the samples with the mixed features of $\bm{c}_{1}$ and $\bm{c}_{2}$.

We verify this idea by conducting the experiments of attribute-to-image~\cite{yan2016attribute2image} on LFW dataset~\cite{huang2008labeled}.
Specifically, it requires us to generate facial images according to the input attributes.
Each image ($\bm{x}_{0}$) in LFW corresponds to a 73-dim real-valued vector ($\bm{c}$), where the value of each dimension represents the degree of some attribute such as male, beard and so on.
We employ Quadratic-Shift with a trainable shift predictor to train on the training set and evaluate it on the test set.
Figure~\ref{fig:lfw1} presents some samples, which shows that ShiftDDPMs can learn a meaningful shift (like a heatmap of the face), and the generated images are consistent with the ground truth in labeled face attributes.
Figure~\ref{fig:lfw2} presents the interpolations generated by Quadratic-Shift.
The interpolations smoothly transition from one side to the other, which verifies our assumptions about the disentangled diffusion trajectories.

\subsection{Image Inpainting}
\begin{figure}[t]
        \centering
        \includegraphics[width=1.0\linewidth]{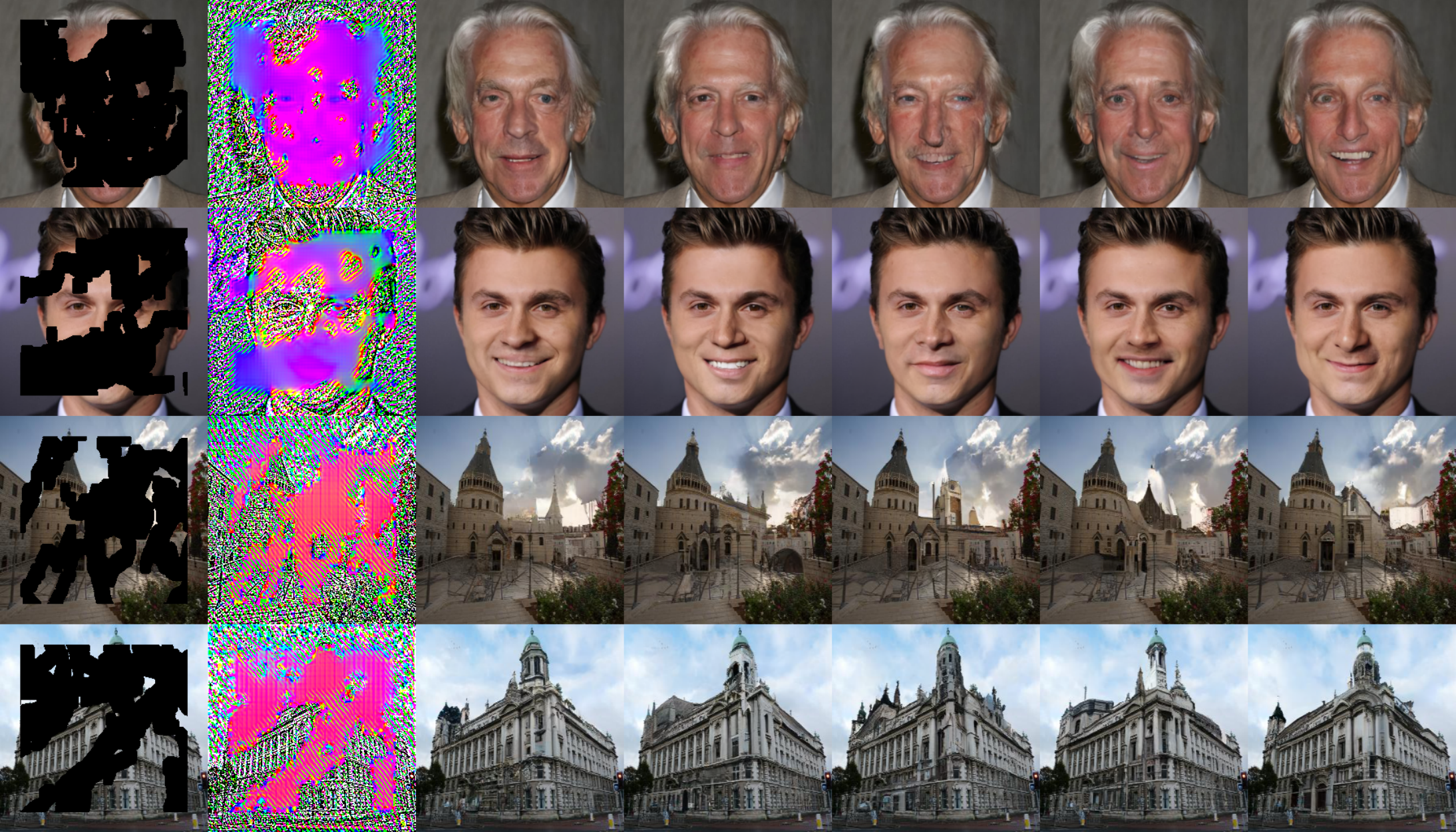}
        \caption{$256 \times 256$ inpainting samples from CelebA-HQ and LSUN-church test set for Quadratic-Shift.}
        \label{fig:inpainting}
\end{figure}

Except for class-conditional image synthesis, we conduct some image-to-image synthesis experiments.
Compared with enumerable class label, image space is almost infinite and it is a challenge to assign a unique trajectory for each instance.
To prove the capacity of ShiftDDPMs, we conduct image inpainting experiments using Irregular Mask Dataset~\cite{liu2018image} with three image datasets: CelebA-HQ~\cite{liu2015deep}, LSUN-church~\cite{yu2015lsun} and Places2~\cite{zhou2017places}.
We employ Quadratic-Shift mode and a UNet based architecture as a shift predictor, which takes as input the masked image and predicts the shift.
Figure~\ref{fig:inpainting} presents some inpainting samples.
As we can see, ShiftDDPMs predict a template of complete image based on the masked one, which guides the trajectory to generate consistent and diverse completions.
To further evaluate ShiftDDPMs on image inpainting, we follow prior works~\cite{yu2019free, liu2018image, zhang2020text} by reporting FID on Places2 dataset.
We choose several GAN-based models: Contextual Attention~\cite{yu2018generative}, EdgeConnect~\cite{nazeri2019edgeconnect} and StructureFlow~\cite{ren2019structureflow} as baselines.
Besides, we take score-based inpainting method proposed in~\cite{song2020score} as another baseline.
Table~\ref{tab:inpainting} presents the quantitative results, and ShiftDDPMs achieve competitive results comparable to prior GAN-based methods.
In addition, ShiftDDPMs also outperform the score-based inpainting method, showing that the extra utilization of the latent space to some extent improves the learning capacity of diffusion models.

\begin{table}
        \centering
        \scalebox{0.9}{
          \begin{tabular}{cccc}
            \toprule
            Mask Percentage & 0-20\% & 20-40\% & 40-60\% \\ 
            \midrule
            Contextual Attention & 4.8586 & 18.4190 & 37.9432 \\
            EdgeConnect & 3.0097 & 7.2635 & 19.0030 \\
            StructureFlow & 2.9420 & 7.0354 & 22.3803 \\
            DDPM (score) & 2.0665 & 6.6129 & 17.3601 \\
            Quadratic-Shift & \bf{1.8314} & \bf{6.2915} & \bf{14.9667} \\
            \bottomrule
          \end{tabular}
        }
    \caption{FID of inpainting results on Places2 dataset.}
    \label{tab:inpainting}
\end{table}

\subsection{Text-to-Image}
\begin{figure}[t]
    \centering
    \includegraphics[width=1.0\linewidth]{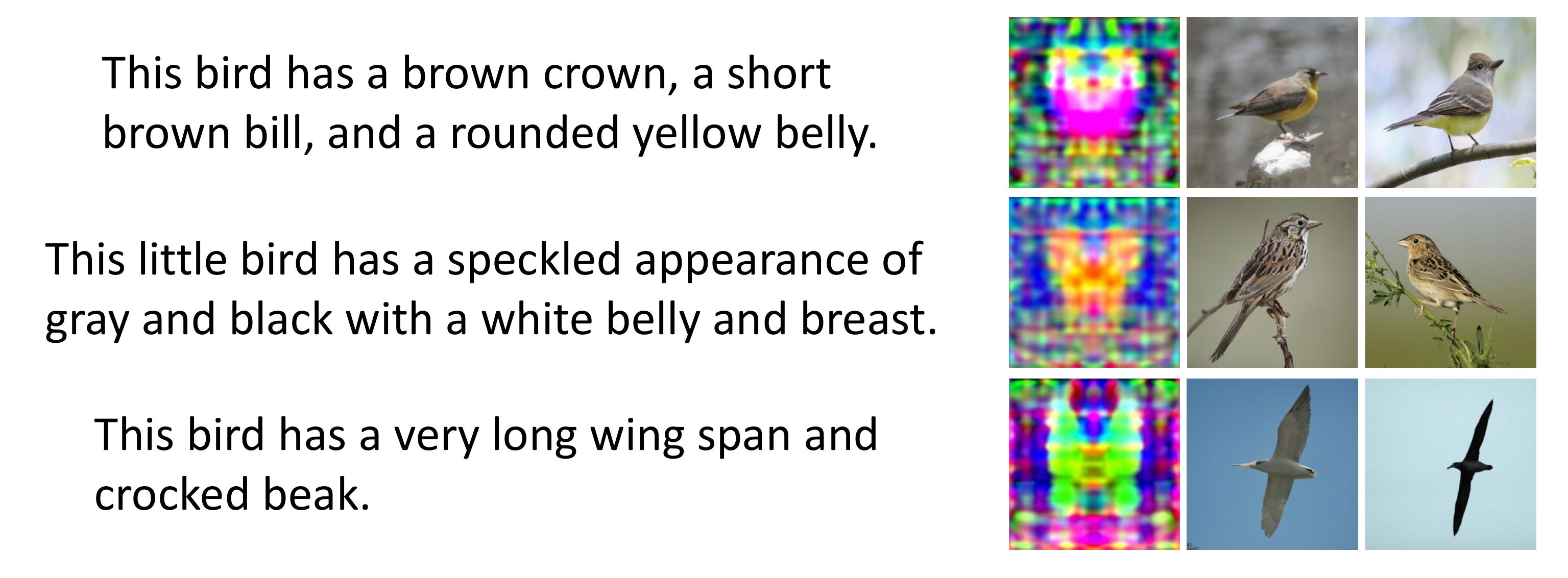}
    \caption{$256 \times 256$ text2img samples from CUB test set for Quadratic-Shift. From left to right are text, learned shift, generated sample and ground truth, respectively.}
    \label{fig:text2img}
\end{figure}

We conduct text-to-image (text2img) experiments on CUB dataset~\cite{wah2011caltech}.
We employ Quadratic-Shift mode and a network as shift predictor to generate shift from the pre-trained sentence embeddings.
Figure~\ref{fig:text2img} presents some generated samples.
We can see that the shift predictor can predict a meaningful template according to text and guide the trajectory to generate text-consistent images.
We choose several GAN-based models GAN-INT-CLS~\cite{reed2016generative}, StackGAN~\cite{zhang2017stackgan}, StackGAN++~\cite{zhang2018stackgan++} and AttnGAN~\cite{xu2018attngan} as baselines.
Besides, we take traditional conditional diffusion method as another baseline, which only incorporates sentence embeddings into the function approximator of parameterized reverse process.
Table~\ref{tab:text2img} presents some quantitative results, and ShiftDDPMs achieve competitive results comparable to prior GAN-based methods and traditional conditional diffusion model.

\subsection{More Choice of $k_{t}$}
The choice of $k_{t}$ is flexible.
For Prior-Shift, any schedules of $k_{t}$ monotonically increasing from $0$ to $1$ can be applied on Prior-Shift.
We have tried with following three types $k_{t}$: $\frac{t}{T}$, $(\frac{t}{T})^{2}$ and $\sin (\frac{t \pi}{2 T} - \frac{\pi}{2})$ and they all work well.
Furthermore, $k_{t}$ can also be piecewise:
\begin{equation}
    \begin{aligned}
    k_{t} = 
        \left\{
            \begin{array}{lr}
            0 & \textrm{$t < 0.4 T$} \\
            \frac{t - 0.4 T}{0.6 T} & \textrm{otherwise}
            \end{array} 
        \right. \,.
    \end{aligned} 
\end{equation}
One can also design other reasonable $k_{t}$.
We leave empirical investigations of $k_{t}$ as future work.

\begin{table}
        \centering
        \scalebox{0.9}{
        \begin{tabular}{ccc}
            \toprule
            Methods & IS & FID \\ 
            \midrule
            GAN-INT-CLS & 2.88 & 68.79 \\
            StackGAN & 3.70 & 51.89 \\
            StackGAN++ & 3.82 & 15.30 \\
            AttnGAN & 4.36 & -\\
            cond. DDPM & 4.18 & 14.79\\
            Quadratic-Shift & \bf{4.42} & \bf{14.26} \\
            \bottomrule
        \end{tabular}
        }
    \caption{IS and FID of text2img results on CUB dataset.}
    \label{tab:text2img}
\end{table}

\section{Conclusion}
In this work, we propose a novel and flexible conditional diffusion model called ShiftDDPMs by introducing conditional forward process with controllable condition-dependent diffusion trajectories.
We analyze the differences of existing related methods under the point of view of ShiftDDPMs and first apply them on image synthesis.
With ShiftDDPMs, we can achieve a better performance and learn some interesting features in latent space.
Extensive qualitative and quantitative experiments on image synthesis demonstrate the feasibility and effectiveness of ShiftDDPMs.

\section{Acknowledgments}
This work was supported in part by the National Natural Science Foundation of China (Grant No.62020106007, No.U21B2040, No.62222211 and No.202100023), Zhejiang Natural Science Foundation (LR19F020006), Zhejiang Electric Power Co., Ltd. Science and Technology Project No.5211YF220006 and Yiwise.

\clearpage
\bibliography{reference}

\clearpage
\appendix

\section*{Appendix A \\ \,}
\section*{Derivation of our conditional forward diffusion kernels}
According to Markovian property, $q(\bm{x}_{t} | \bm{x}_{t-1}, \bm{x}_{0}, \bm{c}) = q(\bm{x}_{t} | \bm{x}_{t-1}, \bm{c})$ for all $t > 1$.
Therefore, we can assume that:
\begin{equation}
    \begin{aligned}
        q(\bm{x}_{t} | \bm{x}_{t-1}, \bm{c}) &= \mathcal{N} (\bm{A}\bm{x}_{t-1} + \bm{b}, \bm{L}^{-1}) \,.
    \end{aligned}
\end{equation}
As we have known the marginal Gaussian for $\bm{x}_{t-1}$:
\begin{equation}
    \begin{aligned}
        q(\bm{x}_{t-1} | \bm{x}_{0}, \bm{c}) = \mathcal{N} (\sqrt{\bar{\alpha}_{t-1}}\bm{x}_{0} + \bm{s}_{t-1} , (1-\bar{\alpha}_{t-1}) \bm{\Sigma}) \,,
    \end{aligned}
\end{equation}
from~\cite{bishop2006pattern} (2.115), we can derive that the marginal Gaussian for $\bm{x}_{t}$, i.e., $q(\bm{x}_{t} | \bm{x}_{0}, \bm{c})$ is given by:
\begin{equation}
    \begin{aligned}
        \mathbb{E}\big[q(\bm{x}_{t} | \bm{x}_{0}, \bm{c})\big] &= \bm{A}(\sqrt{\bar{\alpha}_{t-1}}\bm{x}_{0} + \bm{s}_{t-1}) + \bm{b} \\
        Cov\big[q(\bm{x}_{t} | \bm{x}_{0}, \bm{c})\big] &= \bm{L}^{-1} + (1-\bar{\alpha}_{t-1})\bm{A}\bm{\Sigma}\bm{A}^{T} \,.
    \end{aligned}
\end{equation}
Then we need to ensure that:
\begin{equation}\label{our_forward_direct_2}
    \begin{aligned}
        q(\bm{x}_{t} | \bm{x}_{0}, \bm{c}) = \mathcal{N} (\sqrt{\bar{\alpha}_{t}}\bm{x}_{0} + \bm{s}_{t} \,, \, (1-\bar{\alpha}_{t}) \bm{\Sigma}) \,,
    \end{aligned}
\end{equation}
from which we can derive that:
\begin{equation}
    \begin{aligned}
        \bm{A} &= \sqrt{\alpha_{t}}\bm{I} \\
        \bm{b} &= \bm{s}_{t} - \sqrt{\alpha_{t}} \bm{s}_{t-1} \\ 
        \bm{L}^{-1} &= \big[1-\bar{\alpha}_{t} - \alpha_{t}(1-\bar{\alpha}_{t-1})\big] \bm{\Sigma} = (1 - \alpha_{t}) \bm{\Sigma} \,.
    \end{aligned}
\end{equation}
Finally, for all $t > 1$ we have:
\begin{equation}\label{our_forward_step_2}
    \begin{aligned}
        q(\bm{x}_{t} | \bm{x}_{t-1}, \bm{c}) = \mathcal{N} (\sqrt{\alpha_{t}}\bm{x}_{t-1} + \bm{s}_{t} - \sqrt{\alpha_{t}}\bm{s}_{t-1}, \beta_{t}\bm{\Sigma}) \,.
    \end{aligned}
\end{equation}

We further consider the case for $t = 1$ from following facts:
\begin{equation}
    \begin{aligned}
        q(\bm{x}_{2} | \bm{x}_{1}, \bm{c}) = \mathcal{N} (\sqrt{\alpha_{2}}\bm{x}_{1} + \bm{s}_{2} - \sqrt{\alpha_{2}}\bm{s}_{1}, \beta_{2}\bm{\Sigma}) \\
        q(\bm{x}_{2} | \bm{x}_{0}, \bm{c}) = \mathcal{N} (\sqrt{\bar{\alpha}_{2}}\bm{x}_{0} + \bm{s}_{2} \,, \, (1-\bar{\alpha}_{2}) \bm{\Sigma}) \,.
    \end{aligned}
\end{equation}
With similar derivation based on~\cite{bishop2006pattern} (2.113), we can get:
\begin{equation}
    \begin{aligned}
        q(\bm{x}_{1} | \bm{x}_{0}, \bm{c}) = \mathcal{N} (\sqrt{\alpha_{1}}\bm{x}_{0} + \bm{s}_{1} \,, \, (1-\alpha_{1}) \bm{\Sigma}) \,,
    \end{aligned}
\end{equation}
which matches Eq.(\ref{our_forward_direct_2}).
Therefore we set $\bm{s}_{0} = \bm{0}$, i.e., $k_{0}=0$ to make Eq.(\ref{our_forward_step_2}) true for $t=1$.

One can also verify this conclusion with the recurrence relation in Eq.(\ref{our_forward_step_2}) by the rule of the sum of normally distributed random variables.

\section*{Derivation of the posterior distributions of our conditional forward frocess}
For all $t > 1$, we can derive $q(\bm{x}_{t-1}|\bm{x}_{t},\bm{x}_{0}, \bm{c})$ by Bayes' rule:
\begin{equation}
    \begin{aligned}
        q(\bm{x}_{t-1}|\bm{x}_{t},\bm{x}_{0}, \bm{c}) = \frac{q(\bm{x}_{t}|\bm{x}_{t-1},\bm{x}_{0},\bm{c})\ q(\bm{x}_{t-1}|\bm{x}_{0},\bm{c})}{q(\bm{x}_{t}|\bm{x}_{0},\bm{c})} \,.
    \end{aligned}
\end{equation}
From~\cite{bishop2006pattern} (2.116 and 2.117), we have that $q(\bm{x}_{t-1}|\bm{x}_{t},\bm{x}_{0}, \bm{c})$ is Gaussian and
\begin{equation}
    \begin{aligned}
        Cov\big[ q(\bm{x}_{t-1}|\bm{x}_{t},\bm{x}_{0}, \bm{c}) \big] &= \bigg\{ \frac{1}{1-\bar{\alpha}_{t-1}}\bm{\Sigma}^{-1} +\frac{\alpha_{t}}{1-\alpha_{t}}\bm{\Sigma}^{-1} \bigg\}^{-1} \\
        &= \frac{1}{\frac{1}{1-\bar{\alpha}_{t-1}}+\frac{\alpha_{t}}{1-\alpha_{t}}} \bm{\Sigma} \\
        &= \frac{(1-\bar{\alpha}_{t-1})(1-\alpha_{t})}{1-\alpha_{t}+\alpha_{t}-\bar{\alpha}_{t}} \bm{\Sigma} \\
        &= \frac{1-\bar{\alpha}_{t-1}}{1-\bar{\alpha}_{t}}\beta_{t} \bm{\Sigma} \,,
    \end{aligned}
\end{equation}
and
\begin{equation}
    \begin{aligned}
        &\mathbb{E}\big[ q(\bm{x}_{t-1}|\bm{x}_{t},\bm{x}_{0}, \bm{c}) \big]  \\
        &= \frac{1-\bar{\alpha}_{t-1}}{1-\bar{\alpha}_{t}}\beta_{t}\bm{\Sigma} \bigg[ \frac{\sqrt{\alpha_{t}}}{\beta_{t}}\bm{\Sigma}^{-1}(\bm{x}_{t} - \bm{s}_{t} - \sqrt{\alpha_{t}}\bm{s}_{t-1}) + \\
        & \qquad \qquad \qquad \quad \frac{1}{1-\bar{\alpha}_{t-1}}\bm{\Sigma}^{-1}(\sqrt{\bar{\alpha}_{t-1}}\bm{x}_{0} + \bm{s}_{t-1}) \bigg] \\
        & = \frac{\sqrt{\bar{\alpha}_{t-1}}\beta_{t}}{1-\bar{\alpha}_{t}}\bm{x}_{0} + \frac{\sqrt{\alpha_{t}}(1-\bar{\alpha}_{t-1})}{1-\bar{\alpha}_{t}}\bm{x}_{t} - \sqrt{\alpha_{t}}\frac{1-\bar{\alpha}_{t-1}}{1-\bar{\alpha}_{t}}\bm{s}_{t} + \bm{s}_{t-1} \,.
    \end{aligned}
\end{equation}

\section*{Derivation of the training objective}
The training objective can be represented as:
\begin{equation}
    \begin{aligned}
        L = \mathbb{E}_{q} \bigg\{ &-\log p_{\theta}(\bm{x}_{0}|\bm{x}_{1}, \bm{c}) + D_{KL}\Big[q(\bm{x}_{T}|\bm{x}_{0}, \bm{c})\parallel p(\bm{x}_{T})\Big] \\
        &+ \sum_{t=2}^{T}D_{KL}\Big[ q(\bm{x}_{t-1}|\bm{x}_{t},\bm{x}_{0}, \bm{c})\parallel p_{\theta}(\bm{x}_{t-1}|\bm{x}_{t}, \bm{c})\Big] \bigg\} \,.
    \end{aligned}
\end{equation}

For the first term, we have:
\begin{equation}
    \begin{aligned}
        p_{\theta}(\bm{x}_{0}|\bm{x}_{1}, \bm{c}) = \mathcal{N} (\frac{1}{\sqrt{\alpha_{1}}}(\bm{x}_{1} - \frac{\beta_{1}}{\sqrt{1-\bar{\alpha}_{1}}}\bm{g}_{\theta}(\bm{x}_{1}, 1)) \,, \, \beta_{1}\bm{\Sigma}) \,.
    \end{aligned}
\end{equation}
Then we can derive the first term by Gaussian probability density function:
\begin{equation}
    \begin{aligned}
        & -\log p_{\theta}(\bm{x}_{0} | \bm{x}_{1}, \bm{c}) = \log \big[ (2\pi)^{\frac{d}{2}}  |\beta_{1}\bm{\Sigma}|^{\frac{1}{2}} \big]\\
        &+ \frac{1}{2} \| \bm{x}_{0} - \frac{1}{\sqrt{\alpha_{1}}}(\bm{x}_{1} - \frac{\beta_{1}}{\sqrt{1-\bar{\alpha}_{1}}}\bm{g}_{\theta}(\bm{x}_{1}, 1))  \|^{2}_{(\beta_{1}\bm{\Sigma})^{-1}} \\
        &= \frac{1}{2}  \log \big[(2\pi \beta_{1})^{d} |\bm{\Sigma}| \big] + \frac{1}{2\alpha_{1}} \| \bm{g}_{\theta}(\bm{x}_{1},1) -\frac{\bm{x}_{1} - \sqrt{\bar{\alpha}_{1}}\bm{x}_{0}}{\sqrt{1-\bar{\alpha}_{1}}} \|^{2}_{\bm{\Sigma}^{-1}} \,,
    \end{aligned}
\end{equation}
where $d$ is the dimension of $\bm{x}$.

For the second term, we have:
\begin{equation}
    \begin{aligned}
        q(\bm{x}_{T} | \bm{x}_{0}, \bm{c}) = \mathcal{N} (\sqrt{\bar{\alpha}_{T}}\bm{x}_{0} + \bm{s}_{T} \,,\, (1-\bar{\alpha}_{T}) \bm{\Sigma}) \,,
    \end{aligned}
\end{equation}
and
\begin{equation}
    \begin{aligned}
        p(\bm{x}_{T}) = \mathcal{N}(\bm{s}_{T}, \bm{\Sigma}) \,.
    \end{aligned}
\end{equation}
Then we can derive the second term by Gaussian Kullback$\-$Leibler divergence:
\begin{equation}
    \begin{aligned}
        D_{KL}\Big[q(\bm{x}_{T}|\bm{x}_{0}, \bm{c})\parallel p(\bm{x}_{T})\Big] &= \\
        \frac{1}{2} \big\{ \log \frac{1}{(1-\bar{\alpha}_{T})^{d}} + d(1-\bar{\alpha}_{T}) &- d +\| \sqrt{\bar{\alpha}_{T}} \bm{x}_{0} \|^{2}_{\bm{\Sigma}^{-1}} \big\} \,.
    \end{aligned}
\end{equation}

For the third term, we have:
\begin{equation}
    \begin{aligned}
        q(&\bm{x}_{t-1} | \bm{x}_{t}, \bm{x}_{0}, \bm{c}) = \mathcal{N} (\frac{\sqrt{\bar{\alpha}_{t-1}}\beta_{t}}{1-\bar{\alpha}_{t}}\bm{x}_{0} + \frac{\sqrt{\alpha_{t}}(1-\bar{\alpha}_{t-1})}{1-\bar{\alpha}_{t}}\bm{x}_{t} \\
        &- \frac{\sqrt{\alpha_{t}}(1-\bar{\alpha}_{t-1})}{1-\bar{\alpha}_{t}}\bm{s}_{t} + \bm{s}_{t-1} \,, \, \frac{1-\bar{\alpha}_{t-1}}{1-\bar{\alpha}_{t}}\beta_{t} \bm{\Sigma}) \,,
    \end{aligned}
\end{equation}
and 
\begin{equation}
    \begin{aligned}
        p_{\theta}(\bm{x}_{t-1} | \bm{x}_{t}, \bm{c}) = \mathcal{N} (\frac{1}{\sqrt{\alpha_{t}}}(\bm{x}_{t} - \frac{\beta_{t}}{\sqrt{1-\bar{\alpha}_{t}}}\bm{g}_{\theta}(\bm{x}_{t}, t)) \\
        - \frac{\sqrt{\alpha_{t}}(1-\bar{\alpha}_{t-1})}{1-\bar{\alpha}_{t}}\bm{s}_{t} + \bm{s}_{t-1}  \,, \, \frac{1-\bar{\alpha}_{t-1}}{1-\bar{\alpha}_{t}}\beta_{t} \bm{\Sigma}) \,.
    \end{aligned}
\end{equation}
Then we can derive the third term by Gaussian Kullback$\-$Leibler divergence:
\begin{equation}
    \begin{aligned}
        & D_{KL}\Big[ q(\bm{x}_{t-1}|\bm{x}_{t},\bm{x}_{0}, \bm{c})\parallel p_{\theta}(\bm{x}_{t-1}|\bm{x}_{t}, \bm{c})\Big]  \\
        &= \frac{1}{2} \| \frac{\beta_{t}}{\sqrt{\alpha_{t}}\sqrt{1-\bar{\alpha}_{t}}}(\bm{g}_{\theta}(\bm{x}_{t},t) - \frac{\bm{x}_{t} - \sqrt{\bar{\alpha}_{t}}\bm{x}_{0}}{\sqrt{1-\bar{\alpha}_{t}}}) \|^{2}_{(\frac{1-\bar{\alpha}_{t-1}}{1-\bar{\alpha}_{t}}\beta_{t} \bm{\Sigma}))^{-1}} \\
        &= \frac{\beta_{t}}{2\alpha_{t}(1-\bar{\alpha}_{t-1})} \| \bm{g}_{\theta}(\bm{x}_{t},t) - \frac{\bm{x}_{t} - \sqrt{\bar{\alpha}_{t}}\bm{x}_{0}}{\sqrt{1-\bar{\alpha}_{t}}}) \|^{2}_{\bm{\Sigma}^{-1}} \,.
    \end{aligned}
\end{equation}

Combining the above derivations, we can get final training objective:
\begin{equation}
    \begin{aligned}
        L = c + \sum_{t=1}^{T} \gamma_{t} \mathbb{E}_{\bm{x}_{0},\epsilon}\bigg[ \| \frac{\bm{x}_{t} - \sqrt{\bar{\alpha}_{t}}\bm{x}_{0}}{\sqrt{1-\bar{\alpha}_{t}}} - \bm{g}_{\theta}(\bm{x}_{t}, t) \|^{2}_{\bm{\Sigma}^{-1}} \bigg] \,,
    \end{aligned}
\end{equation}
where $c$ is some constant, $\bm{x}_{0} \sim q(\bm{x_{0}})$, $\bm{\epsilon} \sim \mathcal{N}(\bm{0},\bm{\Sigma})$, $\bm{x}_{t} = \sqrt{\bar{\alpha}_{t}}\bm{x}_{0} + \bm{s}_{t} + \sqrt{1-\bar{\alpha}_{t}}\bm{\epsilon}$, $\|\bm{x}\|^{2}_{\bm{\Sigma}^{-1}} = \bm{x}^{T}\bm{\Sigma}^{-1}\bm{x}$, $\gamma_{1}=\frac{1}{2\alpha_{1}}$ and $\gamma_{t}=\frac{\beta_{t}}{2\alpha_{t}(1-\bar{\alpha}_{t-1})}$ for $t \ge 2$.

\section*{A discretization of Grad-TTS}
Grad-TTS defines a forward process with following SDE:
\begin{equation}
    \begin{aligned}
        \mathrm{d} \bm{X}_{t} = \frac{1}{2} (\bm{\mu} - \bm{X}_{t})\beta_{t}\mathrm{d}t + \sqrt{\beta_{t}}\mathrm{d}\bm{W}_{t} \,,
    \end{aligned}
\end{equation}
where $\bm{\mu}$ corresponds to $\bm{E}(\bm{c})$ of our notations.
Consider a discretization of it:
\begin{equation}
    \begin{aligned}
        \bm{X}_{t+1} -\bm{X}_{t} &= \frac{1}{2} (\bm{\mu} - \bm{X}_{t})\beta_{t} \Delta t + \sqrt{\beta_{t}}\bm{z}_{t} \\
        \bm{X}_{t+1} &= (1 - \frac{1}{2}\beta_{t}\Delta t)\bm{X}_{t} + \frac{1}{2}\beta_{t}\Delta t \bm{\mu} + \sqrt{\beta_{t}}\bm{z}_{t} \\
        &= (1 - \frac{1}{2}\beta_{t}\Delta t)\bm{X}_{t} + (1 - 1 + \frac{1}{2}\beta_{t}\Delta t)\bm{\mu} + \sqrt{\beta_{t}}\bm{z}_{t} \\
        &\approx \sqrt{1-\beta_{t}\Delta t} \bm{X}_{t} + (1 - \sqrt{1-\beta_{t}\Delta t}) \bm{\mu} + \sqrt{\beta_{t}}\bm{z}_{t} \\
        &= \sqrt{\alpha_{t}} \bm{X}_{t} + (1 - \sqrt{\alpha_{t}}) \bm{\mu} + \sqrt{1 - \alpha_{t}}\bm{z}_{t} \,,
    \end{aligned}
\end{equation}
where $\bm{z}_{t} \sim \mathcal{N} (\bm{0}, \bm{I})$ because for Wiener process $W_{t} - W_{s} \sim \mathcal{N} (0, t-s)$ when $0 \le s \le t$.
With this recurrence relation, we can derive that:
\begin{equation}
    \begin{aligned}
        \bm{X}_{t} = \sqrt{\bar{\alpha}_{t}} \bm{X}_{0} + (1 - \sqrt{\bar{\alpha}_{t}}) \bm{\mu} + \sqrt{1 - \bar{\alpha}_{t}} \bm{\epsilon} \,,
    \end{aligned}
\end{equation}
where we can get $k_{t} = 1 - \sqrt{\bar{\alpha}_{t}}$ for Grad-TTS.

\section*{\, \\ Appendix B \\ \,}
\section*{Implementation Details}
We use the same settings with ADM~\cite{dhariwal2021diffusion}, including network architecture, timesteps, variance schedule, dropout, learning rate and EMA.
We set batch size to $128$ for CIFAR-10, $64$ for LFW and $32$ for the others.
We use $4$ feature map resolutions for $32\times32$ models and $6$ for the others.

To compute $\bm{E}_{\psi}(\bm{c})$, we employ a linear layer and stacked transposed convolution layers to map conditions (one-hot vector or attribute vector) to three-channel feature maps for CIFAR-10 and LFW dataset.
For image inpainting on CelebA-HQ, LSUN-church, Place2 datasets, we employ a U-Net architecture for pixel-to-pixel prediction.
For text-to-image synthesis on CUB bird dataset, we employ a linear layer and stacked transposed convolution layers with attention mechanism to map the pre-trained word embeddings to three-channel feature maps.

For image inpainting, we use Irregular Mask Dataset collected by~\cite{liu2018image}, which contains 55,116 irregular raw masks for training and 24,866 for testing. During training, for each image in the batch, we first randomly sample a mask from 55,116 training masks, then perform some random augmentations on the mask, finally we use it to mask the image and get our class center for training. So the training masks are different all the time. The mask is irregular and may be 100\% hole due to augmentations.
During testing, we use 12,000 test masks sampled and augmented from 24,866 raw testing masks. These 12,000 masks are categorized by hole size according to hole-to-image area ratios~(0-20\%, 20-40\%, 40-60\%).

The classifier for (cls. DDPM) employs the encoder half UNet to classify the noisy images.
For the class-conditional function approximator, we use AdaGN same with that in ADM~\cite{dhariwal2021diffusion}.

We train all our models on eight Nvidia RTX 2080Ti GPUs.

\end{document}